# Data-driven multinomial random forest

Junhao Chen[1], Xueli wang[1]

[1]Faculty of Science, Beijing University of Technology, Beijing, China

chenjunhao@emails.bjut.edu.cn

## Abstract

In this article, we strengthen the proof methods of some previously weakly consistent variants of random forests into strongly consistent proof methods, and improve the data utilization of these variants, in order to obtain better theoretical properties and experimental performance. In addition, based on the multinomial random forest (MRF) and Bernoulli random forest (BRF), we propose a data-driven multinomial random forest (DMRF) algorithm, which has lower complexity than MRF and higher complexity than BRF while satisfying strong consistency. It has better performance in classification and regression problems than previous RF variants that only satisfy weak consistency, and in most cases even surpasses standard random forest. To the best of our knowledge, DMRF is currently the most excellent strongly consistent RF variant with low algorithm complexity.

## 1. Introduction

Random Forest (RF, also called standard RF or BriemanRF)[1] is an ensemble learning algorithm that makes classification or regression predictions by taking the majority vote or average of the results of multiple decision trees. Due to its simple and easy-to-understand nature, rapid training, and good performance, it is widely used in many fields, such as data mining[2,3,4], computer vision[5,6,7], ecology[8,9], and bioinformatics[10].

Although the RF algorithm has excellent performance in practical problems, analyzing its theoretical properties is quite difficult due to its highly data-dependent tree-building process. These theoretical properties include consistency, which can be weak or strong. Weak consistency refers to the expectation of the algorithm's loss function converges to the minimum value as the data size tends to infinity, while strong consistency refers to the algorithm's loss function itself converges to the minimum value as the data size tends to infinity[11]. Consistency is an important criterion for evaluating whether an algorithm is excellent, especially in the era of big data.

Many people have made important contributions to the discussion of consistency-related issues in RF, proposing many variants of RF with weak consistency, such as Denil14 (also called Poisson RF)[12], Bernoulli RF (BRF)[13], and Multinomial RF (MRF)[14]. However, the common feature of these algorithms is that the selection of split points and the determination of the final leaf node labels during the tree-building process are independent, i.e., using half of the training set samples to train the split points and the remaining half to determine the leaf node labels, which to a large extent causes insufficient growth of the basic decision trees. In addition, the introduction of random distributions such as Poisson distribution, Bernoulli distribution, and multinomial distribution in these algorithms can enhance their robustness but also have a certain impact on their overall performance.

In this article, we propose a new variant of random forest algorithm called the Data-driven Multinomial Random Forest (DMRF) that has strong consistency, based on the foundation of MRF and BRF with weak consistency. The term "data-driven" in this context does not mean that other variants of random forest algorithm are not data-driven, but rather indicates that the DMRF algorithm can make more effective use of data compared to the aforementioned variants with weak consistency. In the DMRF algorithm, we incorporate a bootstrap method (slightly different from the standard bootstrap method in BriemanRF) that was not included in the previous variants, and introduce a Bernoulli distribution when splitting nodes. This Bernoulli distribution determines whether to select a feature subspace using an optimal splitting criterion to obtain the splitting feature and splitting value, or to randomly extract the splitting feature and splitting feature value using two multinomial distributions based on impurity reduction. The reason for introducing the multinomial distribution is that it can perform random sampling of the optimal splitting feature and feature value with maximum probability[14].

## 2. Related work

BriemanRF[1] is an ensemble algorithm based on the prediction results of multiple decision trees, proposed by Brieman. It has shown satisfactory performance in practical applications. The basic process of BriemanRF can be divided into three steps: first, use the bootstrap method to resample the dataset for the same number of times as the size of the dataset to obtain the training set for the basic decision tree;

second, randomly sample a feature subspace of size $\sqrt{D}$ without replacement from the entire feature space of size $D$, and evaluate the importance of each feature and feature value in the subspace based on the reduction in impurity (e.g., Information entropy or Gini index) to obtain the optimal splitting feature and splitting value, which form the splitting point of the node. Recursively repeat this process until the stopping condition is met, and a decision tree is obtained. Finally, repeat the above process to train multiple decision trees, and take the majority vote (for classification problems) or average (for regression problems) of the results of multiple decision trees to obtain the final prediction result.

Since the proposal of the RF model, many variants have been developed, such as Rotation Forest[15], Quantile RF[16], Skewed RF[17], Deep Forest[18], and Neural RF[19]. These variants are proposed to further enhance the interpretability, performance, and efficiency of the random forest. Although the practical research on RF has developed rapidly, the progress of its theoretical exploration is slightly lagging behind. Brieman proved that the performance of BriemanRF is jointly determined by the correlation between basic trees and the performance of basic trees, i.e., the smaller the correlation between trees (the greater the diversity), and stronger the trees (the better the performance), the better the performance of RF[1].

The important breakthrough in the study of consistency of RF was proposed by Biau et al.[20]. Biau proposed two simplified algorithms of BriemanRF: Purely Random Forest and Scale-Invariant Random Forest. Purely Random Forest randomly selects a feature and its feature value as the splitting feature and splitting value at each node. Scale-Invariant Random Forest also randomly selects a feature as the splitting feature at each node and randomly divides the samples into two parts according to the order of the feature values of that feature. Biau proved that both of these simplified versions have consistency.

Biau et al.[21] proved another simplified RF model that is closer to BriemanRF and has weak consistency: randomly selecting a feature subspace at each node, and for each candidate feature, selecting the midpoint as the splitting value. When selecting among candidate features, the splitting feature and splitting value with the maximum reduction in impurity are selected to grow the tree.

Denil et al.[22] proposed a new RF variant, Denil14 model, which is closer to RF. Denil14 divide the training set into a structural part and an estimation part. The structural part is used only for training the split points, and the estimation part is used only to determine the labels of the leaf nodes. In addition, at each node, the size of the feature subspace is selected based on the Poisson distribution, and the optimal splitting feature and splitting value are searched from m structural part samples that have been pre-selected. This variant has been proven to have weak consistency. Denil14 can be used for classification.

Inspired by the Denil14 model, Wang Yisen et al. proposed a new Bernoulli RF (BRF) based on the Bernoulli distribution[13]. Similar to Denil14, BRF divides the dataset into a structural part and an estimation part, trains the split points using the structural part, and determines the leaf node labels using the estimation part. However, BRF introduces two Bernoulli distributions at each node: one to determine whether the feature subspace size is 1 or $\sqrt{D}$, and the other to determine whether the splitting value of each candidate feature is randomly selected or using the optimal splitting criterion. Finally, they proved that BRF also has weak consistency, but it is closer to BriemanRF and has better performance than previous RF variants with weak consistency. BRF can be used for both classification and regression.

Bai Jiawang et al. [14] transformed the reduction of each feature and its impurity value into probabilities through the softmax function, proposing a Multinomial Random Forest (MRF). MRF also divides the dataset into structural and estimation parts, using the structural part to train the splitting point and the estimation part to determine the leaf node label. Training the splitting point primarily involves two steps: first, calculating the maximum impurity reduction for each feature at each node and converting it into a probability, which is then considered the probability of the multinomial distribution from which the splitting feature is randomly selected; second, when selecting the splitting value, converting the impurity reduction of each feature value of the split feature obtained in the previous step into a probability and regarding it as the probability of the multinomial distribution from which the splitting value is randomly selected. Regarding determining the leaf node label, MRF views the proportion of each class of estimation part samples in the leaf node as a probability and randomly selects a class as the label from the multinomial distribution. MRF has more purposeful random selection of splitting features and values, with more reasonable probability allocations. It currently has the best performance among the consistent variants of RF, even surpassing BriemanRF. The disadvantage is its high computational complexity and large computational cost. MRF is only used for classification.

Due to the fact that Denil14, BRF, and MRF all chose to make the training of split points and leaf node labels independent to achieve weak consistency, this inevitably affects the performance of the base tree and reduces the overall performance of the algorithm. In addition, introducing too much randomness when splitting nodes may increase the robustness of the model, but it will also greatly affect its performance. Based on this, we propose the Data-driven Multinomial Random Forest (DMRF) algorithm that can be used for both classification and regression problems. This algorithm can directly select the sample for the training of the split point to determine the leaf node label. Moreover, this algorithm introduces bootstrap to increase the diversity between trees. More importantly, we enhance the weak consistency to strong consistency by modifying the conditions of the variants mentioned above. We found that although the theoretical basis is different, the proof methods are the same as before. This means that the method for proving weak consistency can be strengthened to the method for proving strong consistency, resulting in a strong conclusion.

  The arrangement of this paper is approximately as follows: Section 3 provides a detailed introduction to the classification DMRF algorithm and proves its strong consistency; Section 4 introduces the regression DMRF algorithm and proves its strong consistency; Section 5 describes the experiment part and introduces the application of the DMRF algorithm in classification and regression problems; and Section 6 concludes the paper.

# 3. Classification DMRF

Let $D_n = \{(X_1, Y_1), (X_2, Y_2), ..., (X_n, Y_n)\}$ denote a dataset where $X_i \in \mathcal{R}^D$ indicates D-dimensional features, $Y_i \in \{1, 2, ..., c\}$ indicates the label, $i \in \{1, 2, ..., n\}$, we are preparing to build $M$ trees.

## 3.1 Training sample sampling

In the DMRF, we use a slightly different bootstrap method than the standard one to sample the training set $D_n^{(j)}$, $j \in \{1, 2, ..., M\}$ for the $j$-th base tree. Specifically, during sampling, we do not resample all samples, but instead sample each sample with a probability of $q_n$ (which may be related to $n$, in this paper, we choose a constant value in $(0,1]$). That is, the sampling of each sample follows a binomial distribution $B(1, q_n)$, and the probabilities $q_n$ among different samples are independent. If the sampled training set is empty, we resample again.

## 3.2 Split point training process

Reviewing the growth process of a standard classification decision tree: Let $V = \{v_{ij}\}$ denote all possible split points for the node $\mathcal{D}$, $v_{ij}$ representing the $j$-th feature's $i$-th feature value (i.e., threshold) in the feature subspace; let $I_{ij}$ denote the impurity reduction obtained by the split point $v_{ij}$ at that node,

$$I_{ij} = I(\mathcal{D}, v_{ij}) = T(\mathcal{D}) - \frac{|\mathcal{D}^l|}{|\mathcal{D}|} T(\mathcal{D}^l) - \frac{|\mathcal{D}^r|}{|\mathcal{D}|} T(\mathcal{D}^r),$$

where $T(\mathcal{D})$ denote impurity criteria of node $\mathcal{D}$, such as Gini index or Information entropy (in this paper, we use the Gini index), are used for measuring impurity, and $\mathcal{D}^l$ and $\mathcal{D}^r$ respectively represent the left and right child nodes obtained by the split at that node.

The impurity reduction obtained by calculating different feature values of feature $A_j$ as split values forms a vector $I^{(j)} = (I_{1,j}, I_{2,j}, ..., I_{m,j})$, $j \in \{1, 2, ..., D\}$, where m represents the number of feature values of feature $A_j$ at that node.

The maximum impurity reduction for each feature forms a vector

$$I = (I_1, I_2, ..., I_D) = (\max I^{(1)}, \max I^{(2)}, ..., \max I^{(D)}).$$

In a standard classification decision tree, the split point is determined by the feature and feature value corresponding to the maximum impurity reduction, i.e., the splitting feature is the $j$-th feature, $j = \arg\max\{I_1, I_2, ..., I_D\}$ and the splitting feature value is the $i$-th feature value of feature $A_j$, $i = \arg\max I^{(j)}$. After splitting the root node into left and right child nodes, the above process is repeated in each child node until the stopping condition is met, at which point the tree stops growing.

Different from the above standard classification tree construction process, DMRF algorithm first conducts a Bernoulli experiment $B$ with probability p when splitting nodes, $B \sim B(1, p)$:

If $B = 1$, the feature subspace with size $\sqrt{D}$ is randomly selected, and the split feature and split value at this node are obtained according to the above optimal split criterion.

If $B = 0$, then the feature subspace is the full feature space, and the split feature and split value at this node can be obtained by the following steps:

① Split feature selection:

i) Normalize $I = (I_1, I_2, ..., I_D) = (\max I^{(1)}, \max I^{(2)}, ..., \max I^{(D)})$ as

$$\tilde{I} = \frac{1}{\max I - \min I}(I_1 - \min I, I_2 - \min I, ..., I_D - \min I);$$

ii) Compute the probablities $\alpha = soft\max(B_1 \tilde{I})$, where $B_1 \geq 0$;

iii) Randomly select a splitting feature according to the multinomial distribution $M(\alpha)$.

② Split feature selection:

Assuming $A_j$ is the split feature which is selected from the previous step.

i) Normalize $I^{(j)} = (I_{1,j}, I_{2,j}, ..., I_{m,j})$ as

$$\tilde{I}^{(j)} = \frac{1}{\max I^{(j)} - \min I^{(j)}}(I_{1,j} - \min I^{(j)}, I_{2,j} - \min I^{(j)}, ..., I_{D,j} - \min I^{(j)});$$

ii) Compute the probablities $\beta = soft\max(B_2 \tilde{I}^{(j)})$, where $B_2 \geq 0$;

iii) Randomly select a splitting value according to the multinomial

distribution $M(\beta)$.

## 3.3 Leaf node label determination

When an unlabeled sample $x$ is given and the prediction is made on it, $x$ will fall on a leaf node of the tree according to the algorithm flow. In the tree, the probability that the $x$ is predicted to be class $k$ is

$$\gamma^{(k)}(x) = \frac{1}{N(\mathcal{N}(x))} \sum_{(X,Y)\in\mathcal{N}(x)} \mathcal{I}(Y=k), k=1,2,...,c,$$

where $\mathcal{I}(\cdot)$ is 1 if $\cdot$ is true and is 0 if $\cdot$ is false; $\mathcal{N}(x)$ is the leaf node where $x$ falls into. According to the majority voting principle, the prediction of sample $x$ under this tree is

$$\hat{y}(x) = \arg\max_{k}\{\gamma^{(k)}(x)\}.$$

The prediction of DMRF is the result of majority voting in the base tree, i.e

$$\bar{y}(x) = \arg\max_{k}\sum_{i=1}^{M}\mathcal{I}(\hat{y}^{(i)}(x)=k),$$

where $\hat{y}^{(i)}(x)$ is the predicted value of sample $x$ in the $i$-th decision tree. If there are multiple categories with the same number of votes, we randomly select one of them as the final prediction class.

The pseudo-code of DMRF algorithm and decision tree construction is as follows:

---
Algorithm1    DMRF classification tree construction process
---

1. Input: A training set for bootstrap sampling $D_n'$, hyper-parameters $p$, $k_n$, $B_1$, $B_2$.
2. Output: A classification tree in DMRF.
3. While $|\mathcal{D}| \geq k_n$:
4.     Compute the impurity reduction of all possible split points $v_{ij}$ at node $\mathcal{D}$.
5.     Do a Bernoulli experiment $B$, $B \sim B(1,p)$;
6.     if $B=1$ then
7.         Select the feature subspace with size $\sqrt{D}$, and use the best split criterion to select the best splitting feature and splitting value.

8.       else

           The vector $I$ composed of the maximum impurity reduction of each feature in the feature full space is normalized to $\tilde{I}$, compute the probability $\alpha = \text{soft}\max(B_1\tilde{I})$, randomly select a splitting feature according to $M(\alpha)$.

9.       The impurity reduction vector of feature $A_j$ selected in the previous step is normalized to $\tilde{I}^{(j)}$, compute the probability $\beta = \text{soft}\max(B_2\tilde{I}^{(j)})$, randomly select the splitting features according to $M(\beta)$.

10.     Split the node $\mathcal{D}$ into left and right child nodes $\mathcal{D}^l, \mathcal{D}^r$ by the split features and values obtained.

11.     if $|\mathcal{D}^l| \geq k_n$ and $|\mathcal{D}^r| \geq k_n$ then

12.         Go to line 3 for $\mathcal{D}^l$ and $\mathcal{D}^r$, recursively build the tree.

13.     else

14.         Meet the stop condition.

15.     end if

16. end while

17. Return: A classification tree in DMRF.

---

Algorithm2    DMRF classification algorithm

1. Input: Training set $D_n = \{(X_1, Y_1), (X_2, Y_2), \ldots, (X_n, Y_n)\}$, the number of trees $M \in N^+$, hyper-parameter $p$, $q_n$, $k_n$, $B_1$, $B_2$, sample $x$.

2. Output: DMRF's prediction for sample $x$.

3. for $i = 1, 2, \ldots, M$ do

4.     Using bootstrap with the probability of $q_n$ to get the training set $D_n^{(i)}$.

5.     if $D_n^{(i)} = \emptyset$ then

6.         Go to line 4 for resampling.

7.     Training a classification tree with the training set $D_n^{(i)}$.

8. end for

9. Return: Predict the class of $x$ by majority voting.

## 3.4 Strong consistency proof of classification DMRF

In this section, we prove the strong consistency of the classification DMRF algorithm, and the detailed proof process is in the appendix.

First, we talk about the consistency definition of classifier. For a classifier sequence $\{g_n\}$, the classifier $g_n$ is obtained by training the data set $D_n = \{(X_1, Y_1), ..., (X_n, Y_n)\}$ which satisfying the distribution $(X, Y)$, and the error rate is

$$L_n = L(g_n) = P(g_n(X, C, D_n) \neq Y \mid D_n),$$

where $C$ is the randomness introduced in the training.

**Definition 3.1**: Given the training set $D_n$ which contain $n$ i.i.d observations, for a certain distribution $(X, Y)$, call classifier $g_n$ is weakly consistent if $g_n$ satisfying

$$\lim_{n \to \infty} EL_n = \lim_{n \to \infty} P(g_n(X, D_n) \neq Y) = L^*,$$

where $L^*$ denotes the Bayes risk. Besides, call $g_n$ is strongly consistent if $g_n$ satisfying

$$P(\lim_{n \to \infty} L_n = L^*) = 1.$$

**Definition 3.2**: A sequence of classifiers $\{g_n\}$ is called weakly(strongly) universally consistent if it is weakly(strongly) consistent for all distributions.

Obviously, the condition of strong consistency is stronger than weak consistency, so strong consistency can derive weak consistency, but vice versa is not necessarily true.

Here are some important lemmas that will be used in the proof.

**Lemma 3.1:** Assume that the classifier sequence $\{g_n\}$ is (universally) strongly consistent, then the majority voting classifier $\overline{g}_n^{(M)}$ (for any value of $M$) is also (universally) strongly consistent.

**Lemma 3.2:** Assume that the classifier sequence $\{g_n\}$ is strongly consistent, the bagging majority voting classifier $\overline{g}_n^M$ (for any value of $M$) is also strongly consistent if $\lim_{n \to \infty} nq_n = \infty$.

Lemma 3.2 is quoted from Theorem 6 [20]. Refer to [20] for more details.

Lemma 3.1 shows that to prove an ensemble classifier with strong consistency, we only need to prove that its base classifier has strong consistency. The universal strong consistency of ensemble classifiers is obtained from the universal strong consistency of base classifiers. Lemma 3.2 can be regarded as a corollary of Lemma 3.1, which shows that the use of bootstrapping does not affect the consistency of the ensemble algorithm. It is worth noting that Lemma 3.1 alone (without bootstrapping) is sufficient to prove the strong consistency of the DMRF. However, using the whole dataset as the training set to build trees will lead to excessive computational costs and high costs with large sample sizes. Moreover, the similarity among the trees will significantly affect the performance of the algorithm. Therefore, we introduce Lemma 3.2, which adds bootstrapping to reduce the training cost while reducing the similarity between trees.

The strong consistency of a single tree is proved below.

**Lemma 3.3:** Let $g_n$ be a binary tree classifier obtained by the split criterion $\pi_n$, whose each region (i.e. leaf node) contains at least $k_n$ points, and $k_n / \log n \to \infty (n \to \infty)$, $A_n(x)$ is the leaf node where the sample $x$ falls into, $\mu(\cdot)$ is the Lebesgue measure of $\cdot$. For all balls $S_r$ with radius $r$ centered at the origin and for all $\gamma > 0$, with probability 1 for all distributions satisfying

$$\lim_{n \to \infty} \mu(\{x : diam(A_n(x) \cap S_r) > \gamma\}) = 0,$$

then $g_n$ corresponding to $\pi_n$ satisfies

$$\lim_{n \to \infty} L(g_n) = L^*$$

with probability one. In other words, $g_n$ is universally strongly consistent.

Lemma 3.3 is quoted from Theorem 21.2[11], refer to [11] for more details. Lemma 3.3 shows that to prove strong consistency of a tree, we only need to prove that any leaf node is small enough, but the sample size in leaf node is large enough.

**Lemma 3.4:** In DMRF, the probability that any feature $A$ chosen to be split feature has lower bound $p_1 > 0$ if $B = 0$ and $B_1$ has upper-bound.

**Lemma 3.5:** Assume that features in DMRF are all supported on $[0,1]$, $B = 0$ and feature $A$ is selected to be the split feature, if this feature is divided into $N(N \geq 3)$ equal partitions $A^{(1)},..., A^{(N)}$ (i.e., $A^{(i)} = [\frac{i-1}{N}, \frac{i}{N}]$), for any split point $v$, $\exists p_2 > 0$, s.t.

$$P(v \in \bigcup_{i=2}^{N-1} A^{(i)} | A) \geq p_2.$$

For detailed proof of the two lemmas, please refer to Lemma 3 and Lemma 4 of [14]. Lemma 3.4 states that when features are randomly selected with multinomial distribution, the probability of each feature being selected is at least a constant $p_1$. Lemma 3.5 states that when a multinomial distribution is used to randomly extract the split value, the probability of the split value falling in the middle part of the interval is at least a constant $p_2$.

Based on the above lemmas, the strong consistency theorem of DMRF algorithm can be obtained:

**Theorem 3.1:** Assume that $X$ is supported on $[0,1]^D$ and has non-zero density almost everywhere, the cumulative distribution function (CDF) of the split points is left-continuous at 1 and right-continuous at 0. If $B_1$, $B_2$ both have upper-bound, DMRF is strongly consistent when $k_n / \log n \to \infty$ and $k_n / n \to 0$ as $n \to \infty$.

## 4. Regression DMRF

In the last section we discussed the DMRF algorithm for classification, in this section we will discuss the DMRF algorithm for regression.

### 4.1 Regression DMRF Algorithm

In classification problems, we choose the Gini index to compute the impurity reduction, while in regression problems, we choose mean squared error(MSE) reduction as the standard for measuring the importance of features and feature values.

Denote the MSE of node $\mathcal{D}$ as

$$MSE(\mathcal{D}) = \frac{1}{N(\mathcal{D})} \sum_{(X,Y) \in \mathcal{D}} (Y - \bar{Y})^2,$$

where $\bar{Y} = \frac{1}{N(\mathcal{D})} \sum_{(X,Y) \in \mathcal{D}} Y$, i.e., the mean of the samples in this node; $N(\mathcal{D})$ is the sample size of node $\mathcal{D}$. Similar to the classification, when the split point is $v_{ij}$, the

MSE reduction is

$$I_{ij} = I(\mathcal{D}, v_{ij}) = MSE(\mathcal{D}) - MSE(\mathcal{D}^l) - MSE(\mathcal{D}^r),$$

where $\mathcal{D}^l$, $\mathcal{D}^r$ denote the left and right child node of $\mathcal{D}$ splitted by $v_{ij}$.

When making prediction, the predicted value of the tree is the sample mean of the leaf node $A_n(x)$ (where sample $x$ resides), in other words,

$$\hat{y}(x) = \frac{\sum_{i=1}^{n} Y_i \mathcal{I}(X_i \in A_n(x))}{\sum_{i=1}^{n} \mathcal{I}(X_i \in A_n(x))} = \frac{1}{N(A_n(x))} \sum_{(X,Y) \in A_n(x)} Y,$$

where $N(A_n(x))$ denotes the sample size of $A_n(x)$. The prediction of the forest is the mean of base trees, that is

$$\bar{y} = \frac{1}{M} \sum_{i=1}^{M} \hat{y}^{(i)}(x),$$

where $M$ denotes the tree number of the forest, $\hat{y}^{(i)}(x)$ is the prediction of $i$-th tree towards $x$.

The difference between the regression DMRF and the classification DMRF lies only in the difference in the splitting criteria for the splitting point and the prediction method. To obtain the regression DMRF, we only need to change the impurity reduction criterion to MSE reduction and the majority voting prediction to mean prediction in the classification DMRF.

## 4.2 Strong consistency proof of Regression DMRF

For a regressor sequence $\{f_n\}$, the regressor $f_n$ is obtained by training the data set $D_n = \{(X_1, Y_1), (X_2, Y_2), ..., (X_n, Y_n)\}$ which satisfying the distribution $(X, Y)$, the MSE of the $f_n$ is

$$R(f_n | D_n) = E[(f_n(X, C, D_n) - f(X))^2 | D_n],$$

where $C$ is the randomness introduced in the training.

Similar to the classification case, let's first define the strong consistency of a regression problem.

**Definition 4.1**: Given the training set $D_n$ which contain n i.i.d observations, for a certain distribution $(X, Y)$, a sequence of regressors $\{f_n\}$ is called weakly consistent if

$f_n$ satisfying

$$\lim_{n\to\infty} E[R(f_n | D_n)] = \lim_{n\to\infty} E[(f_n(X,C,D_n) - f(X))^2] = 0,$$

where $f(X) = E[Y | X]$; $\{f_n\}$ is called is strongly consistent if $f_n$ satisfying

$$\lim_{n\to\infty} R(f_n | D_n) = \lim_{n\to\infty} E[(f_n(X,C,D_n) - f(X))^2 | D_n] = 0$$

with probability one.

**Definition 4.2**: A sequence of regressors $\{f_n\}$ is called weakly (strongly) universally consistent if it is weakly (strongly) consistent for all distributions of $(X,Y)$ with $EY^2 < \infty$.

**Lemma 4.1:** Assume that the regressor sequence $\{f_n\}$ is (universally) strongly consistent, then the averaged regressor $\overline{f}_n^{(M)}$ (for any value of $M$) is also (universally) strongly consistent.

**Lemma 4.2:** Assume that the regressor sequence $\{f_n\}$ is strongly consistent, the bagging averaged regressor $\overline{f}_n^{(M)}$ (for any value of $M$) is also strongly consistent if $\lim_{n\to\infty} nq_n = \infty$.

Lemma 4.1 states that if we want to prove a regression ensemble has strong consistency, we only need to prove that its base regressors have strong consistency. Lemma 4.2 is a corollary of Lemma 4.1 and is similar to the classification case. Bootstrapping is not theoretically necessary but is introduced to reduce computational costs and improve algorithm performance. To prove the consistency of the regression algorithm, Lemma 4.1 is sufficient.

**Lemma 4.3**: Assume for any sphere $S$ centered at the origin

$$\lim_{n\to\infty} \max_{A_{n,j} \cap S \neq \varnothing} diam(A_{n,j}) = 0$$

and

$$\lim_{n\to\infty} \frac{|\{j: A_{n,j} \cap S \neq \varnothing\}| \log n}{n} = 0,$$

then the regressor

$$m'_n = \begin{cases} \dfrac{\sum_{i=1}^{n} Y_i \mathcal{I}(X_i \in A_n(x))}{\sum_{i=1}^{n} \mathcal{I}(X_i \in A_n(x))}, & \sum_{i=1}^{n} \mathcal{I}(X_i \in A_n(x)) > \log n \\ 0, & \text{otherwise} \end{cases}$$

is strongly universally consistent.

Lemma 4.3 is quoted from Theorem 23.2 [11], one can refer to [11] for more details.

**Theorem 4.1**: Assume that $X$ is supported on $[0,1]^D$ and has non-zero density almost everywhere, the cumulative distribution function (CDF) of the split points is left-continuous at 1 and right-continuous at 0. If $B_1$, $B_2$ both have upper-bound, DMRF is strongly consistent when $k_n / \log n \to \infty$ and $k_n / n \to 0$ as $n \to \infty$.

## 5. Experiment

The experiment in this section tests the performance of the proposed DMRF algorithm in the classification and regression problems, and compares it with the other three RF variants with weak consistency and BriemanRF to demonstrate the performance of the DMRF algorithm.

### 5.1 Dataset selection

Our data sets are all from UCI database. Table 1 and Table 2 contain the sample number and feature number of classification and regression data sets respectively, and the classification data set also contains the number of class. In the two tables, we sort the data sets according to the sample number and test the datasets which cover wide range of sample size and feature dimensions in order to show the performance of DMRF. In addition, for missing values of all data sets, we used "-1" padding operation, and no other preprocessing was performed.

Table 1 The description of benchmark classification datasets

| Data sets | Samples | Features | Classes |
|---|---|---|---|
| Blogger | 100 | 6 | 2 |
| Bone marrow | 187 | 39 | 2 |
| Algerian Forest Fires | 244 | 12 | 2 |
| Vertebral | 310 | 6 | 3 |
| Chronic kidney disease | 400 | 25 | 2 |
| Cvr | 435 | 16 | 2 |
| House-votes | 453 | 16 | 2 |
| Wdbc | 569 | 39 | 2 |
| Breast original | 699 | 10 | 2 |
| Balance scale | 625 | 4 | 3 |
| Transfusion | 748 | 5 | 2 |
| Raisin | 900 | 8 | 2 |
| Vehicle | 946 | 18 | 4 |
| Tic-tac-toe | 958 | 9 | 2 |
| Maternal Health Risk | 1014 | 7 | 3 |
| Banknote | 1372 | 4 | 2 |
| Winequality(red) | 1599 | 11 | 7 |
| Car | 1728 | 6 | 4 |
| Wireless | 2000 | 7 | 4 |
| Obesity | 2111 | 17 | 7 |
| Statlog | 2310 | 19 | 7 |
| Ad | 3279 | 1558 | 2 |
| Spambase | 4601 | 57 | 2 |
| Winequality(white) | 4898 | 11 | 7 |
| Page blocks | 5473 | 10 | 5 |
| MFCCs | 7195 | 22 | 4 |
| Mushroom | 8124 | 22 | 7 |
| Electrical Grid | 10000 | 14 | 2 |
| Adult | 48842 | 14 | 2 |
| Connect-4 | 67557 | 42 | 3 |

Table 2  The description of benchmark regression datasets

| Datasets | Samples | Features |
|---|---|---|
| Slump | 103 | 10 |
| ALE | 107 | 6 |
| Alcohol | 125 | 8 |
| Servo | 167 | 4 |
| Computer | 209 | 9 |
| CSM | 217 | 12 |
| Autompg | 398 | 8 |
| Real estate | 414 | 7 |

| Las Vegas Strip | 504 | 20 |
| Housing | 506 | 14 |
| ISTANBUL STOCK | 536 | 8 |
| Qsar fish toxicity | 908 | 7 |
| Concrete | 1030 | 9 |
| QSAR BCF Kow | 1056 | 7 |
| Flare | 1389 | 10 |
| Winequality(red) | 1599 | 11 |
| Communities | 1994 | 128 |
| Skillcraft | 3395 | 20 |
| SML | 4137 | 24 |
| Winequality(white) | 4898 | 11 |
| Parkinsons | 5875 | 26 |
| SeoulBikeData | 8760 | 14 |
| Insurance | 9000 | 86 |
| Combined | 9568 | 4 |
| Cbm | 11934 | 16 |

Note: Due to the large size of the CSM and SeoulBikeData datasets, the labels are log-transformed.

## 5.2 Baselines

We choose three proposed RF variants with weak consistency, Denil14, BRF and MRF, as the comparison model of DMRF. Their common feature is that the dataset is divided into the structure part and the estimation part according to the hyper parameter Ratio, the structure part is used for split points training and the estimation part for leaf node labels determination.

1) Denil14 randomly selects m points of the structure part at each node, then selects the feature subspace with the size of $\min(1+Poisson(\lambda), D)$ without replacing, searches for the optimal splitting point within the range defined by the m points preselected (not the entire number of data points).

2) BRF introduced the first Bernoulli distribution when selecting feature subspace, that is, a feature was randomly selected from the feature set with the probability of $p_1$ as the split feature, or $\sqrt{D}$ features were randomly selected from the feature set with the probability of $1-p_1$ as the candidate feature. The second Bernoulli distribution is introduced in the selection of split values, that is, a value is randomly selected as the split value from the split features with the probability of $p_2$, or the value with the probability of $1-p_2$ is selected from the split features with the largest impurity reduction.

3) MRF normalize the vector composed of the maximum impurity reduction of each feature when selecting the splitting feature, and convert it into probabilities using softmax function, which is used as multinomial distribution to randomly select the splitting feature. The impurity reduction form a vector corresponding to each value of the obtained splitting feature, normalize and convert this vector into probabilities by softmax function, which is used as multinomial distribution to randomly select the splitting eigenvalue.

According to our method, we can abandon the separation of the structural part and the estimation part in the three models mentioned above. At the same time, we can add the bootstrapping method defined earlier. The experiments will examine the performance of these three models in improving data utilization and strong consistency.

## 5.3 Experimental Settings

In the above models, we set $R$atio=0.5 uniformly, minimum sample size of leaf nodes $k_n = 5$, the number of trees $M = 100$. In Denil14, we set $m = 100$, $\lambda = 0.5$ according to [12]. Following [13], we set $p_1 = p_2 = 0.05$. As [14] suggested, in MRF, $B_1 = B_2 = 5$. In DMRF, we choose $q_n = 1 - 1/e$, $p = 0.5$.

## 5.4 Performance Analysis

In the table, "(SE)" indicates that the algorithm is from the original paper, and "(b)" indicates that the algorithm has been rewritten using the bootstrap method defined in this paper and without separating the structural part and the estimation part. In the RF variant, the best performing result in the table is highlighted in bold. To compare the performance of DMRF and BriemanRF, a Wilcoxon signed-rank test with a significance level of 0.05 was used in the experiment. "*" indicates that DMRF is significantly better than BriemanRF, and "#" indicates that BriemanRF is significantly better than DMRF.

### 5.4.1 Classification

The evaluation standard of classification problem is accuracy.

Table 3 Accuracy(%) of different RFs on benchmark datasets

| Datasets | DMRF | MRF(SE) | MRF(b) | BRF(SE) | BRF(b) | Denil14(SE) | Denil14(b) | BriemanRF |
|---|---|---|---|---|---|---|---|---|
| Blogger | **81.89*** | 75.1 | 79.7 | 76.49 | 81.2 | 75.1 | 80.1 | 73.5 |
| Bone marrow | **94.17** | 93.49 | 93.06 | 94.06 | 93.99 | 93.79 | 93.7 | 93.8 |
| Algerian Forest Fires | **93.08** | 92.51 | 92.37 | 92.98 | 93.05 | 92.46 | 92.11 | 93.21 |
| Vertebral | **84.23** | 82.48 | 82.77 | 83.54 | 83.87 | 81.87 | 82.8 | 83.87 |
| Chronic kidney disease | **98.97*** | 98.14 | 98.79 | 98.25 | 98.8 | 98.02 | 98.69 | 98.45 |
| Cvr | **96.04*** | 95.51 | 95.75 | 95.49 | 95.88 | 95.45 | 95.81 | 95.36 |
| House-votes | **96.04*** | 95.5 | 95.98 | 95.58 | 95.94 | 95.45 | 95.82 | 95.45 |
| Wdbc | 96.20 * | 95.78 | **96.27** | 95.36 | 95.92 | 92.86 | 95.77 | 95.71 |
| Breast original | 95.99 | 95.29 | 95.75 | 95.73 | 95.83 | 94.01 | 94.72 | 96.67# |
| Balance scale | 84.98 | 80.31 | 80.94 | 82.48 | 84.82 | 80.69 | 80.14 | 87.14# |
| Transfusion | 76.71 | 73.24 | 75.35 | **78.11** | 76.63 | 77.67 | 74.97 | 78.52# |
| Raisin | **86.13** | 85.76 | 85.95 | 85.69 | 86.01 | 85.61 | 85.94 | 85.89 |
| Vehicle | **75.16*** | 73.54 | 74.87 | 71.67 | 74.85 | 72.81 | 74.55 | 73.52 |
| Tic-tac-toe | **98.73*** | 98.01 | 98.24 | 79.64 | 98.09 | 97.84 | 98.55 | 87.75 |
| Maternal Health Risk | **83.49*** | 77.77 | 83.4 | 77.03 | 82.84 | 77.8 | 82.99 | 76.95 |
| Banknote | 99.39* | 99.49 | **99.69** | 98.32 | 99.31 | 98.29 | 99.09 | 99.06 |
| Winequality(red) | **70.33*** | 63.07 | 70.16 | 63.25 | 70.31 | 63.08 | 69.9 | 66.69 |
| Car | 95.89* | 96.3 | 96.3 | 93.43 | 95.54 | 88.02 | **97.02** | 95.52 |
| Wireless | **98.48*** | 98.06 | 98.3 | 98.02 | 98.14 | 97.67 | 98.11 | 98.15 |
| Obesity | 78.37 | 25.5 | 77.12 | 71.57 | **78.61** | 72.64 | 77.17 | 92.67# |
| Statlog | 98* | **98.35** | 98.25 | 96.87 | 97.77 | 97.41 | 96.77 | 97.09 |
| Ad | 97.25* | 96.76 | **97.95** | 94.43 | 97.46 | 94.16 | 96.98 | 97.02 |
| Spambase | **95.18*** | 93.6 | 95.03 | 93.93 | 95.02 | 91.48 | 95.1 | 94.36 |
| Winequality(white) | **69.56*** | 60.56 | 69.17 | 56.68 | 69.44 | 60.6 | 68.71 | 64.75 |
| Page blocks | **97.59*** | 97.44 | **97.59** | 97.17 | 97.45 | 97.28 | 97.44 | 97.39 |
| MFCCs | 98.47* | 98.02 | **98.5** | 98.02 | 98.47 | 97.83 | 98.36 | 98.2 |
| Mushroom | 56.36* | 52.34 | 47.31 | **60.82** | 57.22 | 51.46 | 47.41 | 48.1 |
| Electrical Grid | 92.41* | 91.46 | **92.52** | 90.7 | 91.94 | 90.38 | 91.45 | 91.76 |
| Adult | **86.45*** | 86.28 | 86.13 | 57.57 | 86.44 | 86.19 | 85.57 | 86.37 |
| Connect-4 | 81.08* | 81.46 | **83.96** | 76.75 | 80.19 | 66.19 | 83.05 | 78.2 |
| Average rank | **1.90** | 5.57 | 3.33 | 5.97 | 3.03 | 6.80 | 4.70 | 4.70 |

The results of the Table 3 indicate that in the majority of cases among RF variants, the DMRF algorithm has the best performance. Additionally, the Denil14, BRF, and MRF variants all obtained performance improvements after being rewritten to strengthen their theoretical properties (for example, all three variants saw accuracy increases of over 8% on the Winequality(white) dataset). The average ranking of the three variants also get a significant improvement, with MRF and BRF both increasing their average ranking by two places. This is notable since the complexity of the variants did not change, indicating that using training samples to determine splitting points and leaf node labels can greatly improve algorithm performance, and these processes should be highly correlated during tree construction.

Considering the comparison between DMRF and BriemanRF, in terms of accuracy, DMRF is significantly better than BriemanRF in most cases. Additionally, DMRF has the highest average ranking, while BriemanRF has a relatively lower ranking. The reason for this is that DMRF has a certain probability of selecting the optimal feature from the entire feature space when determining the splitting points during training, which expands its search range compared to BriemanRF, resulting in higher performance.

Based on the above analysis, it can be inferred that choosing DMRF is the optimal choice for classification problems.

5.4.2 Regression

The evaluation criterion of regression problem is mean square error.

Table 4 Mean square error(%) of different RFs on benchmark datasets

| Datasets | DMRF | MRF(SE) | MRF(b) | BRF(SE) | BRF(b) | Denil14(SE) | Denil14(b) | BriemanRF |
|---|---|---|---|---|---|---|---|---|
| Slump | **29.92** | 53.23 | 32.14 | 59.46 | 40.66 | 62.01 | 48.15 | 15.47# |
| ALE | 0.0578 | 0.0714 | **0.049** | 0.0923 | 0.0603 | 0.0971 | 0.0662 | 0.0508# |
| Alcohol | **0.0172** | 0.0313 | 0.0194 | 0.1166 | 0.0199 | 0.1193 | 0.0303 | 0.0528 |
| Servo | 0.5394 | 0.439 | **0.2791** | 0.6861 | 0.5849 | 0.8742 | 0.6729 | 0.4215# |
| Computer | 2647.1* | 6429.9 | 2992.3 | 13561 | 2985.1 | 13156.6 | **2547.1** | 5569.1 |
| CSM | **30.36*** | 36.41 | 30.46 | 31.13 | 32.21 | 31.32 | 38.33 | 30.95 |
| Autompg | 13.66 | 25.09 | **11.45** | 23.88 | 16.7 | 24.55 | 18.52 | 8.07# |
| Real estate | **54.65** | 126.97 | 54.97 | 120.19 | 60.33 | 110.26 | 93.9 | 55.36 |
| Las Vegas Strip | **0.9901** | 1.075 | 0.9914 | 1.012 | 1.0028 | 1.0229 | 1.0713 | 0.953# |
| Housing | 13.08* | 28.99 | **10.62** | 65.13 | 21.25 | 56.45 | 22.31 | 14.1 |
| ISTANBUL STOCK | **2.16E-04*** | 3.88E-04 | 2.20E-04 | 4.15E-04 | 4.14E-04 | 4.40E-04 | 3.33E-04 | 2.19E-04 |
| Qsar fish toxicity | 0.7764* | 1.5125 | **0.7566** | 2.0155 | 1.1808 | 2.7172 | 1.5126 | 0.8112 |
| Concrete | 39.32 | 144.99 | **27.4** | 256.68 | 41.52 | 260.32 | 44.55 | 31.37# |
| QSAR BCF Kow | 0.7485 | 1.6764 | **0.747** | 1.4094 | 0.9091 | 1.0513 | 1.0479 | 0.5477# |
| Flare | **0.2239** | 0.2457 | 0.2487 | 0.2246 | 0.2243 | 0.2342 | 0.2347 | 0.2222# |
| Winequality(red) | 0.3424* | 0.4703 | **0.3379** | 0.6448 | 0.3651 | 0.6379 | 0.4137 | 0.3482 |
| Communities | **0.0203** | 0.0372 | 0.0205 | 0.0521 | 0.2049 | 0.0499 | 0.0244 | 0.0182# |
| Skillcraft | 4.436 | 6.544 | **4.355** | 6.948 | 5.618 | 7.021 | 6.029 | 4.272# |
| SML | 49.04 | 36.18 | **2.49** | 127.54 | 97.91 | 117.62 | 117.56 | 2.42# |
| Winequality(white) | 0.3605* | 0.549 | **0.3584** | 0.7747 | 0.3767 | 0.7297 | 0.3999 | 0.3953 |
| Parkinsons | **0.0013** | 0.0029 | 0.0014 | 61.61 | 0.0016 | 0.0045 | 0.0016 | 0.0012# |
| SeoulBikeData | 0.9386 | 0.9635 | **0.2061** | 1.4459 | 1.1743 | 1.3006 | 1.3002 | 0.1781# |
| Insurance | **0.0556** | 0.0562 | 0.0587 | 0.06 | 0.056 | 0.056 | 0.056 | 0.0554# |
| Combined | 11.96 | 21.63 | **11.81** | 48.75 | 12.62 | 26.15 | 14.47 | 11.71# |
| Cbm | **3.529** | 1.901 | 6.717 | 5.522 | 5.168 | 5.606 | 5.602 | 7.443* |
| Average Rank | **2.28** | 5.72 | 2.64 | 6.88 | 4.2 | 6.84 | 5.12 | 2.32 |

Compared to classification, the advantages of DMRF in regression problems are not as significant. In the Table 4, DMRF has the best performance on half of the datasets, while MRF(b) is generally the best on the rest. After being rewritten to strengthen their theoretical properties, the Denil14, BRF, and MRF variants all obtained significant performance improvements (for example, the MSE improvement ratio of Computer and Concrete datasets was over 53%), and their rankings also saw a notable improvement, particularly MRF(b), which improved its average ranking by over three places. Furthermore, upon closer examination of the results, it is worth noting that DMRF have good performance on datasets with large amounts of data and numerous features. This is because larger amounts of data result in more comprehensive training, while more features introduce additional randomness that increases the diversity of base decision trees. Therefore, when dealing with regression problems on datasets with large amounts of data and numerous features, DMRF is a reliable choice.

Comparing DMRF and BriemanRF, it is clear that BriemanRF is significantly better in most cases, particularly when sample sizes are large.

Based on the above analysis, it can be concluded that in regression problems, if strong consistency is considered, then DMRF is the optimal choice; otherwise, BriemanRF is a better option.

### 5.5 Parameter Analysis

In this section, we explore the influence of hyper parameters on DMRF. We focus on $p$, $q_n$, $B_1$, $B_2$. For $p$ and $q_n$, the test range we take is [0.05, 0.95] with step size 0.1. For $B_1$, $B_2$ the test range is the integer in $[0, 20]$. In terms of dataset size, we define datasets with less than 500 data points as small, datasets with 500-1000 data points as medium, and datasets with more than 1000 data points as large. For classification problems, accuracy is used as the evaluation metric. For regression problems, negative mean squared error (NMSE) is used as the evaluation metric for ease of observation.

5.5.1 The effect of $p$, $q_n$

We investigate $p$, $q_n$ under $B_1 = B_2 = 10$ as [14] recommended.

1）classification

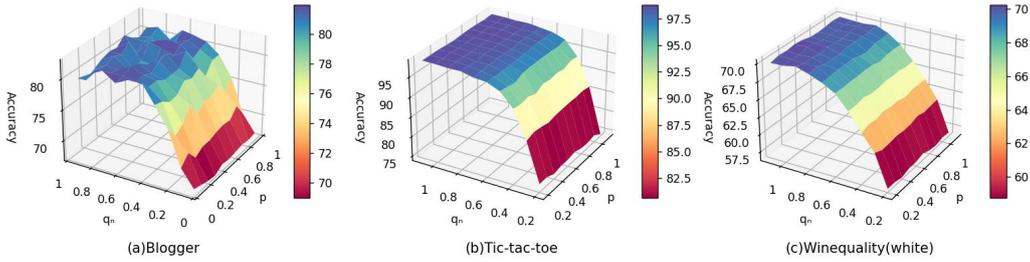

Fig 1. Accuracy(%) of the DMRF under different p, q values

Fig 1 shows the performance of DMRF on three classification datasets with small, medium, and large sample sizes under $B_1 = B_2 = 5$. It can be observed that for the same $p$ value, the accuracy of the three datasets at $q_n = 0.6$ has two situations: close to maximum and stable or decreasing. Through analysis, it is known that when $q_n$ is too small, the sampling probability of each sample is too low, resulting in insufficient training of base trees; as $q_n$ increases, the number of sampled samples increases, and the number of training samples of the base trees gradually becomes sufficient, resulting in the overall performance of the algorithm increasing; However, when the number of samples reaches a certain value, the performance improvement slows down, and there will be a situation where the accuracy tends to be stable or even starts to decrease. The reason for the decrease is that the number of samples taken exceeds an appropriate value, resulting in high similarity between the training sets of the base trees, which affects the overall performance. Since the optimal value of $q_n = 0.6$ is around 0.6, we set $q_n = 1 - 1/e (\approx 0.6322)$ to reduce computational burden while obtaining the optimal parameter.

It is worth noting that the bootstrap method used in this paper can be considered as a non-repeated sampling version of the standard bootstrap method under large samples. In fact, if we take n non-repeated samples of an n-sample dataset with equal probability, then the probability of each sample being selected is $1 - (1 - 1/n)^n \to 1 - 1/e (n \to \infty)$. This is also one of the reasons why we chose the $q_n = 1 - 1/e$.

Under the same $q_n$ value, it can be observed that the accuracy increases first and then decreases with the increase of $p$, and reaches the optimal value at around $p = 0.5$. In the DMRF algorithm, if $p = 0$, the DMRF algorithm is similar to the MRF algorithm, and the selection of split points at each node depends on the random selection of two multinomial distributions; if $p = 1$, each node selects the optimal split point from the feature subspace, which is similar to BriemanRF. When $p$ is small, the probability of selecting split points according to the reduction of impurity at each node results in a higher probability of selecting more important split points, which leads to high similarity between the trees and insufficient overall algorithm performance. As $p$ increases, the probability of selecting the feature subspace increases, selecting the optimal split point in the feature subspace not only increases diversity but also ensures higher performance, so the overall performance increases. When $p$ approaches 1, each node is more likely to select the feature subspace and less likely to consider the full feature space. Although the diversity is greater, it is difficult to select the optimal split point in the entire feature space, resulting in a decrease in overall algorithm performance. In addition, it can be concluded that the algorithm is more sensitive to the parameter $q_n$ and less sensitive to the parameter $p$. This indicates that training sample size is more important than the splitting method.

2）regression

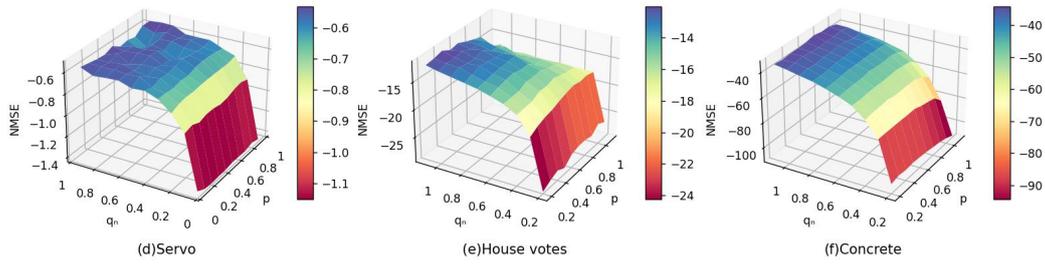

Fig 2. Negative mean square error of the DMRF under different p, q values

Fig 2 shows the performance of DMRF on three regression datasets with small, medium, and large sample sizes under $B_1 = B_2 = 5$. It can be observed that the regression situation is similar to the classification: under the same $p$ value, the NMSE increases with the increase of $q_n$ and reaches its maximum at around 0.6 before stabilizing or beginning to decrease; under the same $q_n$ value, the NMSE initially increases with the increase of $p$, but then decreases or stabilizes after reaching a certain point. Therefore, the optimal $q_n$ value is considered to be $1-1/e$ and the optimal $p$ value is considered to be 0.5. Additionally, it's still observed that the DMRF algorithm is more sensitive to parameter $q_n$ than to parameter $p$.

5.5.2 The effect of $B_1$, $B_2$

1）classification

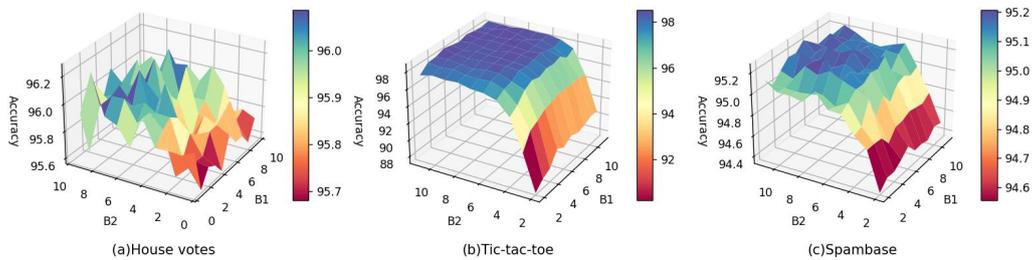

Fig 3. Accuracy(%) of the DMRF under different B1, B2 values

Fig 3 shows the impact of $B_1$ and $B_2$ on the DMRF algorithm under $q_n = 1-1/e$

and $p = 0.5$ for three classification datasets with small, medium, and large sample sizes. Under the same $B_2$ value, the accuracy increases slightly as $B_1$ increases from 0, reaching a stable point at around $B_1 = 5$ before slowly increasing or decreasing. Under the same $B_1$ value, the accuracy starts to increase as $B_2$ increases from 0 and stabilizes at around $B_2 = 5$. The reason is that when $B_1$ is close to 0, the probabilities of each feature are not significantly different from each other, making it difficult to sample the optimal features. As $B_1$ grows, the differences in probabilities between features become larger, and more important features tend to be selected, improving the algorithm's performance. However, when $B_1$ grows to a certain extent, the selection of features becomes similar to selecting the optimal feature, leading to the similarity between base decision trees being too high and causing a decrease in performance. The situation for $B_2$ is similar to that of $B_1$. This conclusion is similar to that in [14].

It can also be seen from the figure that the DMRF is not sensitive to $B_1$ but is more sensitive to $B_2$. Since $B_1$ affects the selection of splitting features and $B_2$ affects the selection of split values, it can be concluded that in classification problems, the choice of split values is more critical than the choice of splitting features.

2）regression

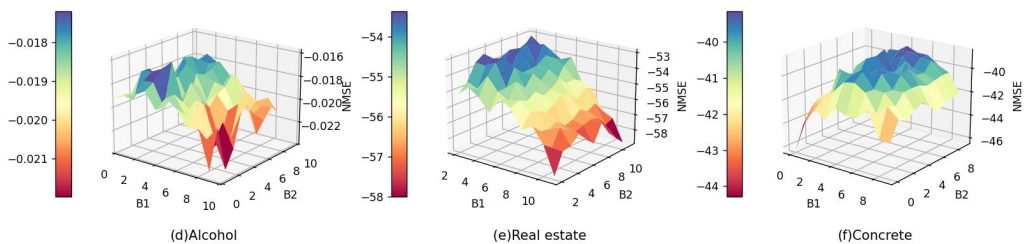

Fig 4 Negative mean square error of the DMRF under different B1, B2 values

Fig 4 shows the impact of $B_1$ and $B_2$ on the DMRF algorithm under $q_n = 1 - 1/e$ and $p = 0.5$ for three regression datasets with small, medium, and large sample sizes. Unlike the classification case, under the same $B_2$ value, the NMSE increases as $B_1$ decreases from 10 in general. After reaching a low point at around $B_1 = 5$, NMSE growth slows or starts to decrease. Under the same $B_1$ value, NMSE starts to increase as $B_2$ increases from 0 and stabilizes or slightly decreases at around $B_2 = 5$. The reason is that when $B_1$ is too large, the probability of selecting the optimal feature is much higher than that of other features, resulting in high similarity between base decision trees and poor performance. As $B_1$ decreases, more randomness is introduced, preserving the high performance of base decision trees while increasing diversity, improving the algorithm's performance. When $B_1$ is close to 0, the probabilities of each feature are not significantly different from each other, making it difficult to sample optimal features, leading to the poor performance of base decision trees and the algorithm.

The situation for $B_2$ is similar to that of $B_1$. However, it can be observed from the figure that the DMRF algorithm is not sensitive to $B_2$ but is more sensitive to $B_1$, which is the opposite of the classification case. Since $B_1$ affects the selection of splitting features and $B_2$ affects the selection of split values, it can be concluded that in regression problems, the choice of splitting features is more critical than the choice of split values.

## 5.6 Computational complexity analysis

Assume that the data set has $n$ samples and $D$ features, we prepare to build $M$ trees. The complexity of random sampling is not considered below.

The best case for tree construction is complete balanced growth, in this case, the depth of the tree is $\mathcal{O}(\log n)$. All samples of each layer of DMRF are involved in the calculation, and the average number of features calculated is $p\sqrt{D} + (1-p)D$. Therefore, the average complexity of building a DMRF tree is $\mathcal{O}([p\sqrt{D} + (1-p)D]n\log n)$, the complexity of DMRF is $\mathcal{O}(nM[p\sqrt{D} + (1-p)D]\log n)$.

In the same way, the complexity of BriemanRF is $\mathcal{O}(nM\sqrt{D}\log n)$. The complexity of Biau08 is $\mathcal{O}(M \log n)$ due to the random selection of split features and split values. The complexity of Biau12 is $\mathcal{O}(M\sqrt{D} \log n)$ [13]. The feature subspace size of Denil14 is $\min(1 + Poisson(\lambda), D)$, and the optimal split point is searched in the pre-selected samples of m structural parts, so its complexity is $\mathcal{O}(\min(1 + Poisson(\lambda), D) \cdot mM \log n)$. BRF introduces two Bernoulli distributions when choosing split points, the average number of features calculated at each layer is $p_1 + (1 - p_1)\sqrt{D}$, and the average number of samples calculated at each layer is $(1 - p_2)n$, so the complexity is $\mathcal{O}((p_1 + (1 - p_1)\sqrt{D})(1 - p_2)nM \log n)$. MRF introduces two multinomial distributions when selecting split points, and all features and samples at each node are involved in the calculation, so the complexity is $\mathcal{O}(nMD \log n)$.

| RF variants | Complexity |
| --- | --- |
| Biau08 | $\mathcal{O}(M \log n)$ |
| Biau12 | $\mathcal{O}(M\sqrt{D} \log n)$ |
| Denil14 | $\mathcal{O}(\min(1 + Poisson(\lambda), D) \cdot mM \log n)$ |
| BRF | $\mathcal{O}((p_1 + (1 - p_1)\sqrt{D})(1 - p_2)nM \log n)$ |
| MRF | $\mathcal{O}(DnM \log n)$ |
| DMRF | $\mathcal{O}([p\sqrt{D} + (1 - p)D]nM \log n)$ |
| BriemanRF | $\mathcal{O}(\sqrt{D}nM \log n)$ |

From the above discussion, it can be seen that MRF has the highest complexity, followed by DMRF. The complexity ranking of BRF, Denil14, and BriemanRF depends on the parameter settings, while Biau12 and Biau08 have the lowest complexity. Therefore, we aim to improve both the theoretical and practical performance of MRF while reducing its complexity.

# 6. Conclusion

The main contributions of this article are as follows:

1. By modifying the previous weak consistency proof method of the RF based on the sample size conditions of the leaf nodes, we improved it to a strong consistency proof method. We also improved the weak consistency proofs of Denil14, BRF, and MRF and achieved strong consistency in classification problems.

2. While BRF has been discussed in both classification and regression cases, Denil14 and MRF have only been discussed in classification cases. In this article, we extended their discussion to regression cases. Among these three variants, the experimental results showed that MRF had the best performance in regression, followed by BRF, and then Denil14.

3. We proposed a DMRF algorithm that combines the Bernoulli distribution and the multinomial distribution. The DMRF algorithm obtains the training set of base decision trees by using a newly defined bootstrap method. During tree construction, it uses the combination of Bernoulli and multinomial distributions to obtain split points, which increases diversity while maintaining high performance of base decision trees. Experimental results show that compare to the previous RF variants with weak consistency proofs, DMRF has higher accuracy.

4. In most cases, DMRF is significantly better than BriemanRF in classification problems, but not as good as BriemanRF in regression problems. This indicates that DMRF is more suitable for classification problems.

5. The experimental part mainly discusses the impact of various parameters of the DMRF algorithm on its performance in both classification and regression problems. The theoretical analysis shows that the complexity of DMRF is between BRF and MRF, and the complexity of Denil14 is the lowest.

The biggest advantage of the DMRF algorithm lies in its good theoretical properties. In addition, DMRF uses the limiting form of the standard bootstrap method to get the training set of base decision trees in large samples. During the tree growth process, it fully utilizes the advantages of accurate feature space searching and diverse feature subspace searching, and determines the label of leaf nodes using the sample at the training split points. Compared to models with weak consistency proofs such as Denil14, BRF, and MRF, DMRF has a higher data utilization rate.

The disadvantage of DMRF is that its complexity is only second to MRF and its computational cost is relatively high compared to Denil14, BRF, and BriemanRF. In addition, its performance in regression problems is average. Therefore, under the premise of not requiring consistency, BriemanRF would be a better choice in terms of cost-effectiveness.

Future work could focus on improving DMRF to reduce its complexity, thereby reducing computational resources and increasing computational efficiency.

# Appendix

**The proof of Lemma 3.1:**

Denote $g^*(x)$ as the Bayes classifier, then the Bayes risk is

$$L^* = P(g^*(x) \neq Y).$$

Denote

$$A = \{k \mid \gamma^{(k)}(x) = \max\{\gamma^{(k)}(x)\}\},$$

$$B = \{k \mid \gamma^{(k)}(x) < \max\{\gamma^{(k)}(x)\}\}.$$

Then

$$P(\bar{g}_n^{(M)}(x,C,D_n) \neq Y \mid D_n)$$

$$= \sum_k P(\bar{g}_n^{(M)}(x,C,D_n) = k \mid D_n) P(Y \neq k \mid D_n)$$

$$\leq L^* \cdot \sum_{k \in A} P(\bar{g}_n^{(M)}(x,C,D_n) = k \mid D_n) + \sum_{k \in B} P(\bar{g}_n^{(M)}(x,C,D_n) = k \mid D_n)$$

so it is sufficient to prove that the limit of the latter term is 0 for all $k \in B$.

For $\forall k \in B$,

$$P(\bar{g}_n^{(M)}(x,C,D_n) = c \mid D_n)$$

$$= P(\sum_{i=1}^M \mathcal{I}(g_n(x,C^{(i)},D_n) = k) > \max_{l \neq k}\{\sum_{i=1}^M \mathcal{I}(g_n(x,C^{(i)},D_n) = l)\} \mid D_n)$$

$$\leq P(\sum_{i=1}^M \mathcal{I}(g_n(x,C^{(i)},D_n) = k) \geq 1 \mid D_n)$$

$$\leq E(\sum_{i=1}^M \mathcal{I}(g_n(x,C^{(i)},D_n) = k) \mid D_n)$$

$$= MP(g_n(x,C,D_n) = k) \mid D_n) \to 0 (n \to \infty).$$

□

**The proof of Lemma 4.1:**

Every base tree is strongly consistent, i.e.,

$$\lim_{n\to\infty} R(f_n | D_n) = \lim_{n\to\infty} E[(f_n(X, C^{(i)}, D_n) - f(X))^2 | D_n] = 0, i \in \{1, 2, ..., M\}.$$

Then

$$R(\overline{f}_n^{(M)} | D_n)$$

$$= E[(\frac{1}{M}\sum_{i=1}^{M} f_n(X, C^{(i)}, D_n) - f(X))^2 | D_n]$$

$$\overset{(c)}{\leq} \frac{1}{M}\sum_{i=1}^{M} E[(f_n(X, C^{(i)}, D_n) - f(X))^2 | D_n] \to 0(n \to \infty).$$

where $C^{(i)}$ is the randomness introduced in $i$-th tree building, (c) uses Cauchy inequality.

□

**The proof of Theorem 3.1:**

It is sufficient to show that it satisfies the conditions of Lemma 3.3. Obviously, we just need to prove that $diam(A_n(x)) \to 0$ as $n \to \infty$. Let $V(i)$ denote the size of the $i$-th feature of $A_n(x)$, we only need to show that $E[V(i)] \to 0$ for all $i \in \{1, 2, ..., D\}$.

For a given $i$, denote the largest size among its child nodes as $V^*(i)$.

Define $C$ as the event that select the feature subspace and obtain the optimal split point by optimal partition, then $\overline{C}$ is the event that obtain split point by two multinomial distribution. Besides, $P(C) = p$, $P(\overline{C}) = 1 - p$.

By lemma 3.4,

$$E[V^*(i) | \overline{C}] \leq (1 - p_2) \cdot 1 + p_2 \frac{N-1}{N} \cdot 1 = 1 - \frac{p_2}{N}.$$

By lemma 3.5,

$$E[V(i) | \overline{C}] \leq (1 - p_1) \cdot 1 + p_1 E[V^*(i) | \overline{C}] \leq 1 - \frac{1}{N} p_1 p_2.$$

Then

$$E[V(i)] = P(C)E[V(i) | C] + P(\overline{C})E[V(i) | \overline{C}]$$

$$\leq p \cdot 1 + (1 - p) \cdot E[V^*(i) | \overline{C}]$$

$$\le p + (1-p)(1 - \frac{1}{N} p_1 p_2) \triangleq A < 1.$$

The above process is the result of one time split. If the $i$-th feature is splited m times, the following formula can be obtained by iterating the above formula continuously:

$$E[V(i)] \le A^m.$$

As long as $m \to \infty (n \to \infty)$ with the probability 1 is proved to be true, that is, each feature can be split infinitely times, the strong consistency of DMRF can be obtained.

Due to the randomness of the split point selection, the final selected split point can be regarded as a random variable W, which follows the uniform distribution on [0,1], and its cumulative distribution function is

$$F_W(x) = x, x \in [0,1].$$

For $\forall m \in N^+, \varepsilon > 0$ and a certain $0 < \eta < 1$, the smallest child node after the root node splits according to a splitting feature is denoted as $M_1 = \min(W, 1-W)$, then we have

$$P(M_1 \ge \eta^{1/m}) = P(\eta^{1/m} \le W \le 1 - \eta^{1/m})$$
$$= F_W(1 - \eta^{1/m}) - F_W(\eta^{1/m})$$
$$= 1 - \eta^{1/m} - \eta^{1/m}$$
$$= 1 - 2\eta^{1/m}.$$

Without loss of generality, we can normalize the value of all attributes to range [0,1] for each node. If the feature is continuously split m times (i.e. the tree grows to the m layer), the probability that the smallest child node in m layer has the size at least $\eta$ is

$$P(M_m \ge \eta) = (1 - 2\eta^{1/m})^m.$$

In this case, if $0 < \eta < \{\frac{1}{2}[1 - (1-\varepsilon)^{\frac{1}{K}}]\}^K$, then

$$P(M_m \ge \eta) = (1 - 2\eta^{1/m})^m > 1 - \varepsilon.$$

The above results are based on the fact that the same feature is selected for each split. In fact, if different features are split at different layers, $P(M_m \geq \eta)$ will be greater, we still have

$$P(M_m \geq \eta) > 1 - \varepsilon.$$

This indicates that the size of each node is $\eta$ in $m$-th layer with the probability at least $1 - \varepsilon$.

Since X has a non-zero density function, each node in the $m$-th layer of the tree has a positive metric with respect to $\mu_X$. Define

$$q = \min_{\mathcal{N}: a\ leaf\ at\ K-th\ level} \mu_X(\mathcal{N}),$$

$q > 0$ because of the fact that the measure of each leaf node is positive and the number of leaf nodes is finite.

The number of samples in the training set is $n$, and the number of samples in the leaf node $\mathcal{N}$ is $Binomial(n,q)$, then

$$P(N(\mathcal{N}) < k_n) = P(N(\mathcal{N}) - nq < k_n - nq)$$

$$\stackrel{(a)}{=} P(|N(\mathcal{N}) - nq| > |k_n - nq|)$$

$$\stackrel{(b)}{\leq} \frac{nq(1-q)}{|k_n - nq|^2}$$

$$= \frac{q(1-q)}{n|\frac{k_n}{n} - q|^2} \to 0(n \to \infty).$$

(a) is based on the fact that $k_n/n \to 0$ as $n \to \infty$, so $k_n - nq < 0$. (b) uses Chebyshev's inequality. This suggests that the probability of reaching the stop condition will converge to 0 as $n \to \infty$, which means that can split infinitely many times with probability 1.

In summary, $diam(A_n(x)) \to 0(n \to \infty)$ with probability 1, DMRF tree is strongly consistent. By lemma 3.2, DMRF algorithm has strong consistency.

□

**The proof of Theorem 4.2:**

By lemma 4.1, the strong consistency of DMRF in regression problem is based on the strong consistency of trees. The following proves the strong consistency of the base regression tree.

By lemma 4.3, if we prove

$$\lim_{n \to \infty} diam(A_n(x)) \to 0$$

and

$$\lim_{n \to \infty} \frac{|\{j : A_{n,j} \cap S \neq \varnothing\}| \log n}{n} = 0,$$

then the strong consistency of

$$m'_n = \begin{cases} \dfrac{\sum_{i=1}^{n} Y_i \mathcal{I}(X_i \in A_n(x))}{\sum_{i=1}^{n} \mathcal{I}(X_i \in A_n(x))}, & \sum_{i=1}^{n} \mathcal{I}(X_i \in A_n(x)) > \log n \\ 0, \text{其他} \end{cases}$$

is obtained. Since the sample number of each leaf node is at least $k_n$, i.e.,

$$N(A_n(x)) = \sum_{i=1}^{n} \mathcal{I}(X_i \in A_n(x)) \geq k_n.$$

From $k_n / \log n \to \infty (n \to \infty)$, when $n$ is sufficiently large,

$$\sum_{i=1}^{n} \mathcal{I}(X_i \in A_n(x)) \geq k_n > \log n.$$

In this case

$$m'_n = \hat{y}(x) = \frac{\sum_{i=1}^{n} Y_i \mathcal{I}(X_i \in A_n(x))}{\sum_{i=1}^{n} \mathcal{I}(X_i \in A_n(x))} = \frac{1}{N(A_n(x))} \sum_{(X,Y) \in A_n(x)} Y.$$

That is, the base regression tree is universally strongly consistent. Therefore, we only need to prove that the conditions of the above two limits are true. The former condition has been proved in the consistency proof of classification DMRF algorithm, so only the latter is needed to prove.

For the base decision tree with n training samples, since the sample size of each leaf node is at least $k_n$ and there are at most $\dfrac{n}{k_n}$ split regions,

$$\frac{|\{j : A_{n,j} \cap S \neq \varnothing\}| \log n}{n} \leq \frac{n}{k_n} \cdot \frac{\log n}{n} = \frac{\log n}{k_n} \to 0 (n \to \infty).$$

The latter is true. Therefore, the strong consistency of the regression DMRF algorithm is obtained.